\documentclass[runningheads,orivec]{llncs}
\usepackage[T1]{fontenc}
\usepackage{newtxtext}       %
\usepackage[varvw]{newtxmath}       

\usepackage{graphicx}

\usepackage{multirow}
\begin{document}
\title{Labadain-MedQA: A Tetun Medical QA Dataset for Fine-Tuning and Evaluating Large Language Models}
%
\titlerunning{Labadain-MedQA: A Tetun Medical QA Dataset}
%
\author{Gabriel de Jesus\inst{1}\orcidID{0000-0003-4392-2382} \and
S{\'e}rgio Nunes\inst{1,2}\orcidID{0000-0002-2693-988X}}
\authorrunning{G. de Jesus and S. Nunes}
%
\institute{INESC TEC, Porto, Portugal\\
\email{gabriel.jesus@inesctec.pt}\\
\and
University of Porto, Porto, Portugal\\
\email{sergio.nunes@fe.up.pt}}
\maketitle              
\begin{abstract}
Advances in large language models (LLMs) have created opportunities for cross-lingual transfer of medical knowledge from high-resource to low-resource languages to support access to healthcare information. However, medical question-answering (QA) datasets for fine-tuning and evaluating LLMs remain scarce for Tetun.
To address this gap, we translated MedAlpaca's Medical Flashcards and WikiDoc Patient Information collections into Tetun using Gemini 2.5 Flash, producing 39,077 medical QA pairs. 
Translation quality was assessed on 50 randomly sampled QA pairs (100 documents) using automatic metrics against human reference translations and evaluations by two native Tetun-speaking healthcare professionals, focusing on medical accuracy and Tetun fluency.
Human translations achieved higher mean ratings than LLM-generated translations for both criteria (4.43/5 vs. 4.02/5), with statistically significant differences (Wilcoxon signed-rank test, $p < 0.001$). Nevertheless, LLM-generated translations also received high ratings.
To assess downstream utility, we fine-tuned Gemma and LLaMA variants on the translated MedAlpaca dataset using parameter-efficient fine-tuning (PEFT) and evaluated them on 325 Tetun-translated United States Medical Licensing Examination multiple-choice questions.
Gemma 4 12B fine-tuned using QLoRA achieved relative improvements in mean accuracy of 10.79\% with 4-bit NormalFloat (NF4) inference and 10.42\% after merging the LoRA adapters into the BF16 base model, compared with the corresponding base models.
These results highlight the viability of automatic dataset translation for Tetun medical QA and the effectiveness of PEFT in enhancing model performance.
The datasets, models, and evaluation resources are accessible at \url{https://github.com/labadain/labadain-medqa} to support reproducible research on Tetun medical QA and facilitate machine translation, cross lingual, and multilingual research.

\end{abstract}

\section{Introduction}

Advances in artificial intelligence (AI), particularly large language models (LLMs), have enabled access to healthcare information through natural-language interaction. However, these advances have been uneven across languages, as most models are developed using large text corpora and training resources predominantly available in high-resource languages. Languages with limited digital resources have therefore received less attention. Tetun, one of the official languages of Timor-Leste~\cite{vasconcelos-et-al-2011}, is among these languages. Despite its widespread use in public communication, education, media, and daily life, Tetun remains underrepresented in natural language processing (NLP) and AI research. 

Conversational AI applications are increasingly used to access information, including health-related topics. For Tetun speakers, Labadain Chat\footnote{\url{https://www.labadain.com}} provides a general-purpose conversational AI application~\cite{jesus-nunes-2026-labadainchat}. An analysis of Labadain Chat interaction logs found that healthcare accounted for 28\% of user queries, making it one of the most frequent topics alongside education and science~\cite{jesus-nunes-2025-insights}. However, generating accurate healthcare responses remains challenging for LLMs in Tetun, where domain-specific training and evaluation resources are scarce.

Cross-lingual transfer offers an opportunity to address this limitation by leveraging resources developed for high-resource languages and adapting them to low-resource ones~\cite{conneau-et-al-2020,agrawal-et-al-2023}. We therefore investigate the feasibility of translating existing English medical QA resources into Tetun and assess their utility for adapting open-weight LLMs through parameter-efficient fine-tuning (PEFT). Our study is guided by the following research questions: \textit{(R1)~How effectively can medical QA resources be transferred from English to Tetun through automatic translation?} and \textit{(R2)~How does PEFT on translated Tetun medical QA data affect LLM performance on downstream tasks?}

\begin{figure}
\includegraphics[width=0.99\textwidth]{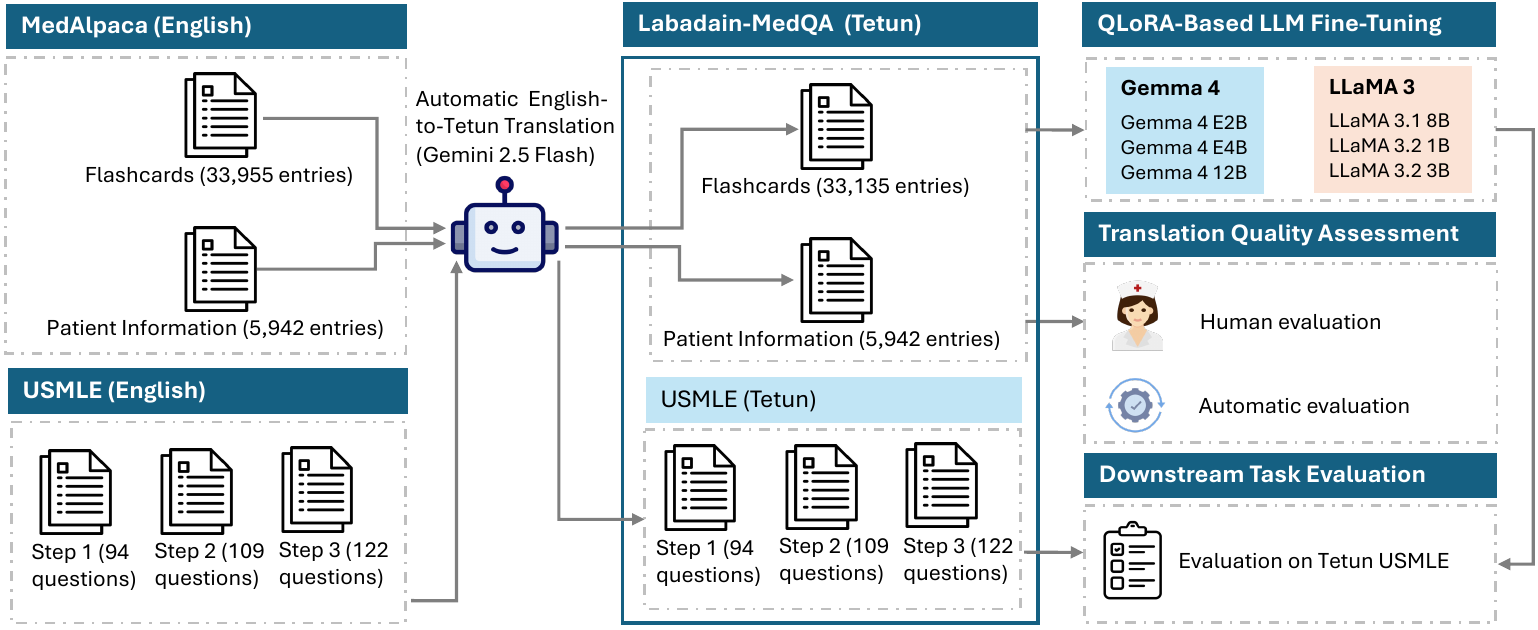}
\Description{Labadain-MedQA workflow for constructing Tetun medical datasets for LLM fine-tuning and evaluation.}
\caption{Overview of the Labadain-MedQA workflow for constructing Tetun medical datasets for LLM fine-tuning and evaluation.}
\label{fig:Labadain-MedQA-workflow}
\end{figure}

To address \textit{R1}, we automatically translated the Medical Flashcards and WikiDoc Patient Information collections from MedAlpaca~\cite{han-et-al-2023} into Tetun using Gemini 2.5 Flash (see Fig.~\ref{fig:Labadain-MedQA-workflow}). A sample of 50 QA pairs (100 individual documents) was randomly selected from the translated dataset, and the corresponding original English texts were provided to a native Tetun-speaking translator for translation into Tetun. The resulting human translations served as references for evaluating the automatic translations. Subsequently, two native Tetun-speaking healthcare professionals assessed the quality of both human and automatically generated translations of the same 100 documents, focusing on medical accuracy and Tetun fluency. Additionally, automatic translation quality was assessed using automatic evaluation metrics computed against the human reference translations for these 100 documents.
For \textit{R2}, we used the translated MedAlpaca dataset to fine-tune models from the Gemma 4 and LLaMA 3 families using Quantized Low-Rank Adaptation (QLoRA)~\cite{dettmers-et-al-2023-qlora}, a PEFT method. We then evaluated and compared the fine-tuned models on USMLE datasets~\cite{usmle-2023} translated into Tetun using the same translation approach as for MedAlpaca, to assess the effectiveness of fine-tuning and establish baselines for Tetun medical QA.

The main contributions of this study are: \textit{(1)~Tetun medical QA dataset:} we introduce \textit{Labadain-MedQA}, comprising 39,077 medical QA pairs from MedAlpaca~\cite{han-et-al-2023} and 325 multiple-choice questions from USMLE, all automatically translated from English into Tetun; \textit{(2)~Translation quality assessment and evaluation datasets:} we provide a systematic evaluation of automatic medical translation into Tetun and release the human reference translations and annotation datasets; \textit{(3)~Fine-tuned models and downstream evaluation:} we demonstrate the utility of the translated dataset through PEFT of open-weight LLMs, establish baselines on a Tetun medical multiple-choice QA task, and release the resulting fine-tuned models to support future research.

\section{Related Work}
The development of medical QA systems relies on domain-specific datasets that provide medical knowledge for training and evaluating language models. These datasets serve different purposes, including medical education and patient information. Several English-language medical QA datasets have been developed, including MedQuAD~\cite{ben-demner-2019}, MedMCQA~\cite{pal-et-al-2022}, and MedAlpaca~\cite{han-et-al-2023}. MedAlpaca introduced Medical Meadow, a collection of medical instruction datasets comprising over 160,000 examples for fine-tuning LLMs. The collection includes Medical Flashcards~\cite{medalpaca-medical-flashcards}, containing 33,955 QA pairs designed for medical education, and WikiDoc Patient Information~\cite{medalpaca-wikidoc-patient-info}, comprising 5,942 QA pairs that provide information about medical conditions and patient care.

To address the limited availability of healthcare resources in LRLs, several studies have developed bilingual and multilingual datasets for medical translation and question answering. For medical translation, Lankford et al.~\cite{lankford-etal-2022} developed gaHealth, an English--Irish bilingual health corpus for domain-specific machine translation. 
Merx et al.~\cite{merx-et-al-2025-openwho} introduced OpenWHO, a parallel corpus of healthcare documents covering more than 20 languages, including Tetun and eight other LRLs, and evaluated LLMs for healthcare translation.
For medical QA, Kimera et al.~\cite{kimera-et-al-2026} introduced a multilingual maternal healthcare dataset comprising 3,694 QA pairs in English, Luganda, Runyankore, and Swahili, providing medical information in languages with limited healthcare resources.

The development of healthcare resources through translation requires careful quality assessment, as errors may alter medical meaning and introduce clinical risks. 
Nigatu et al.~\cite{nigatu-et-al-2025} investigated machine translation errors in Amharic and Tigrinya, identifying challenges involving medical terminology, procedural descriptions, and linguistic differences. 
To improve translation quality, Merx et al.~\cite{merx-etal-2025-tulun} introduced TULUN, combining neural machine translation with glossary-guided LLM post-editing to improve medical translation into Tetun.
Kimera et al.~\cite{kimera-et-al-2026} employed forward and back translation, followed by validation by medical professionals to assess clinical accuracy. 
These studies highlight the importance of translation quality control and human evaluation when developing medical language resources for underrepresented languages.

The utility of medical resources can be further assessed by using them to fine-tune language models and evaluating their performance on downstream tasks. Given the computational costs of full fine-tuning, parameter-efficient fine-tuning (PEFT) methods offer more efficient approaches to adapting LLMs to specialized domains. Low-Rank Adaptation (LoRA)~\cite{hu-et-al-2022-lora} reduces the number of trainable parameters, while Quantized Low-Rank Adaptation (QLoRA)~\cite{dettmers-et-al-2023-qlora} further reduces memory requirements through 4-bit quantization. Unsloth~\cite{han-han-2023-unsloth} provides an optimized framework for implementing LoRA and QLoRA, enabling faster fine-tuning with reduced memory consumption.
 
For LRLs, Bui et al.~\cite{bui-et-al-2025} employed LoRA and QLoRA to fine-tune LLaMA 2 (7B and 13B) and BLOOMZ 3B models using Vietnamese medical data and reported improvements over the corresponding base models on downstream text generation tasks. Alhanai et al.~\cite{alhanai-et-al-2025} fine-tuned LLaMA 3 70B using translated Winogrande and Massive Multitask Language Understanding (MMLU) college medicine datasets across eight low-resource African languages and evaluated its performance on multiple-choice reasoning, clinical knowledge, virology, and reading comprehension tasks. More recently, Ulli and Mondal~\cite{ulli-mondal-2025-medqwen} employed QLoRA to fine-tune Qwen3-1.7B using multilingual medical dialogue datasets across ten languages and evaluated its performance on question answering, summarization, and information extraction tasks. These studies provide a basis for adapting LLMs to medical QA for LRLs such as Tetun.

For Tetun, previous research has primarily focused on general-purpose language resources and applications. Labadain-30k+~\cite{de-jesus-nunes-2024-labadain30k+} introduced a manually audited Tetun document collection for NLP and information retrieval, while Labadain Chat~\cite{jesus-nunes-2026-labadainchat} demonstrated the application of LLMs to general-purpose conversational interactions in Tetun. Building on these efforts, we introduce Labadain-MedQA, a Tetun medical QA dataset comprising Medical Flashcards and WikiDoc Patient Information collections translated into Tetun for fine-tuning LLMs, alongside USMLE multiple-choice questions translated into Tetun for evaluation. This work addresses the limited availability of domain-specific Tetun resources for developing and evaluating medical QA systems.

\section{The Labadain-MedQA Dataset}

To develop \textit{Labadain-MedQA}, we selected two collections from MedAlpaca covering medical and clinical knowledge relevant to students, healthcare professionals, and the general public, alongside USMLE-based multiple-choice questions for downstream evaluation. These English-language resources were automatically translated into Tetun using Gemini 2.5 Flash. The following subsections describe the data sources, translation process, and characteristics of the resulting dataset.

\subsection{Data Sources}

The following three English-language data sources were selected to construct the fine-tuning and evaluation components of \textit{Labadain-MedQA}.

\paragraph{Medical Flashcards Dataset.} This dataset derives from Anki Medical Curriculum flashcards\footnote{\url{https://apps.ankiweb.net}} developed for medical education, covering subjects such as anatomy, physiology, pathology, pharmacology, and clinical medicine. In MedAlpaca~\cite{han-et-al-2023}, the original flashcards were converted into QA pairs using GPT-3.5 Turbo, resulting in 33,955 pairs with relatively short QA focused on specific medical concepts~\cite{medalpaca-medical-flashcards}.

\paragraph{WikiDoc Patient Information Dataset.} This dataset contains 5,942 medical QA pairs~\cite{medalpaca-wikidoc-patient-info} extracted from WikiDoc,\footnote{\url{https://www.wikidoc.org}} a collaborative medical knowledge platform. Unlike Medical Flashcards, the content was already structured as questions and answers covering medical conditions, symptoms, causes, risk factors, treatment, and prognosis, with explanatory answers intended for patient education.

\paragraph{USMLE Dataset.} This dataset comprises 325 multiple-choice questions drawn from datasets based on the three steps of the United States Medical Licensing Examination (USMLE)~\cite{usmle-2023}. Step 1~\cite{augtoma-usmle-step1} contains 94 questions focusing on foundational medical sciences, Step 2~\cite{augtoma-usmle-step2} contains 109 questions on clinical knowledge and patient care, and Step 3~\cite{augtoma-usmle-step3} contains 122 questions on clinical decision-making and patient management.

\subsection{Translation Pipeline}

\begin{figure}
\includegraphics[width=0.99\textwidth]{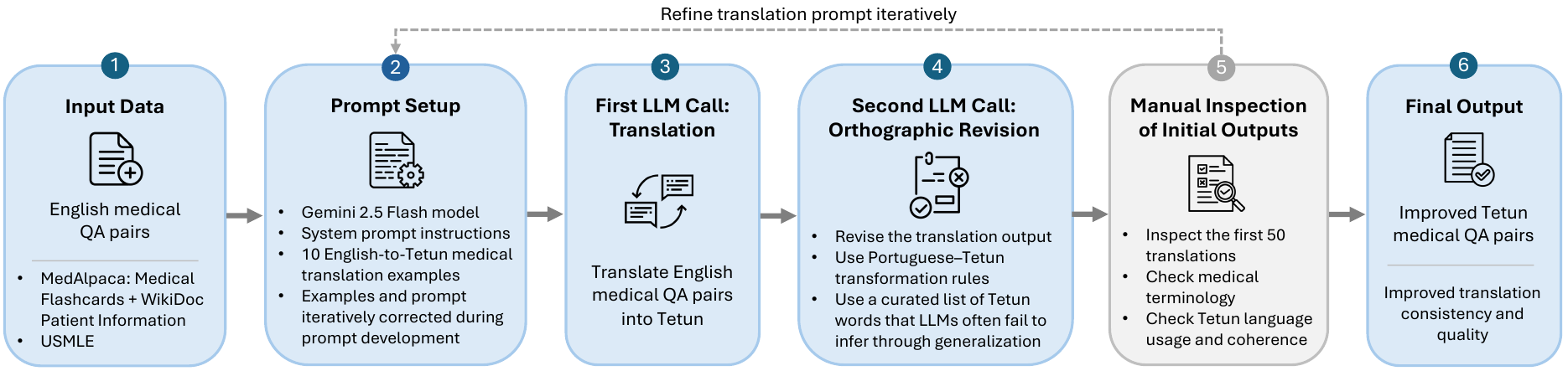}
\Description{Iterative translation and orthography revision pipeline for Labadain-MedQA.Two-stage LLM workflow: translation followed by orthographic revision, with iterative prompt refinement based on manual inspection.}
\caption{Iterative translation and orthography revision pipeline for Labadain-MedQA.}
\label{fig:iterative-translation-pipeline}
\end{figure}

The translation pipeline used to construct \textit{Labadain-MedQA} is illustrated in Fig.~\ref{fig:iterative-translation-pipeline}. It consists of two successive LLM calls for translation and orthographic revision, followed by manual inspection and iterative prompt refinement.

The English medical QA pairs were automatically translated into Tetun using Gemini 2.5 Flash, selected based on its performance in previous Tetun-related work~\cite{jesus-nunes-2026-labadainchat}. We used a prompt structure similar to that used in Labadain Chat~\cite{jesus-nunes-2026-labadainchat}, but with different settings. In the first LLM call, the model was instructed to translate the input text using ten English-to-Tetun medical QA translation examples, and the resulting translations were passed to a second call to the same LLM for orthographic revision, guided by Portuguese--Tetun transformation rules~\cite{standard-tetun-inl-2004} and a curated list of Tetun words that LLMs often fail to infer through generalization. Both prompts were iteratively refined during development.

We adopted an iterative prompting approach using LangSmith\footnote{\url{https://smith.langchain.com}} to trace and inspect the translation and revision outputs from the two-stage process. The first ten outputs were manually inspected for issues in medical terminology, coherence, and Tetun usage, with the inspection progressively extended to 50 documents. These observations guided iterative refinement of both prompts to improve translation consistency and quality.

\subsection{Dataset Composition and Characteristics}

The \textit{Labadain-MedQA} dataset comprises two components: MedAlpaca-derived datasets for LLM fine-tuning and a USMLE-based dataset for evaluation. The MedAlpaca-derived datasets contain 39,077 medical QA pairs from two collections: 33,135 from Medical Flashcards and 5,942 from WikiDoc Patient Information. The original Medical Flashcards collection contained 33,955 pairs, but 820 were excluded due to missing values or parsing errors during translation.
The USMLE dataset contains 325 multiple-choice questions and each question contains answer-option fields labeled A through I, with up to eight populated options. Most questions contain five answer choices, while unused fields contain null values. These questions were used exclusively for medical QA evaluation and were not included in the fine-tuning data.

\begin{figure}
\centering
\setlength{\fboxsep}{0pt}
\setlength{\fboxrule}{0.5pt}
\fcolorbox{gray!50}{white}{
    \includegraphics[width=0.98\textwidth]{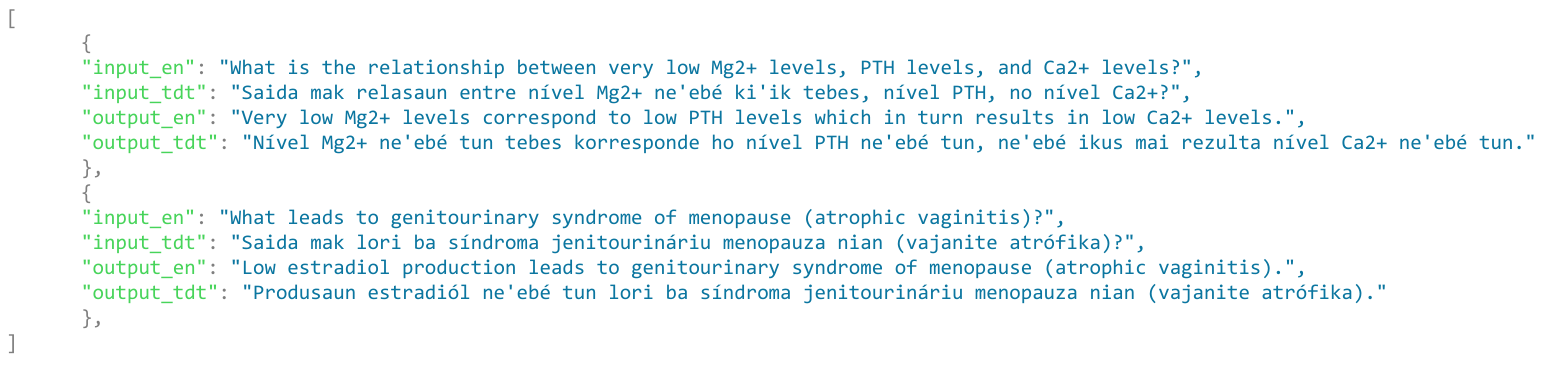}
}
\Description{Examples of the \textit{Labadain-MedQA} dataset structure, showing English source texts and their Tetun translations.}
\caption{Examples of the \textit{Labadain-MedQA} dataset structure, showing English source texts and their corresponding Tetun translations.}
\label{fig:labadain-medqa-dataset-examples}
\end{figure}

The dataset retains the original English texts alongside their corresponding Tetun translations to support future multilingual and cross-lingual research. As illustrated in Fig.~\ref{fig:labadain-medqa-dataset-examples}, each QA pair in the MedAlpaca-derived dataset contains four fields: \texttt{input\_en} and \texttt{input\_tdt} for questions in English and Tetun, respectively, and \texttt{output\_en} and \texttt{output\_tdt} for their corresponding answers. Further details on the datasets are available in the \textit{Labadain-MedQA} repository at \url{https://github.com/labadain/labadain-medqa}.

\begin{figure}
\centering
\includegraphics[width=0.98\textwidth]{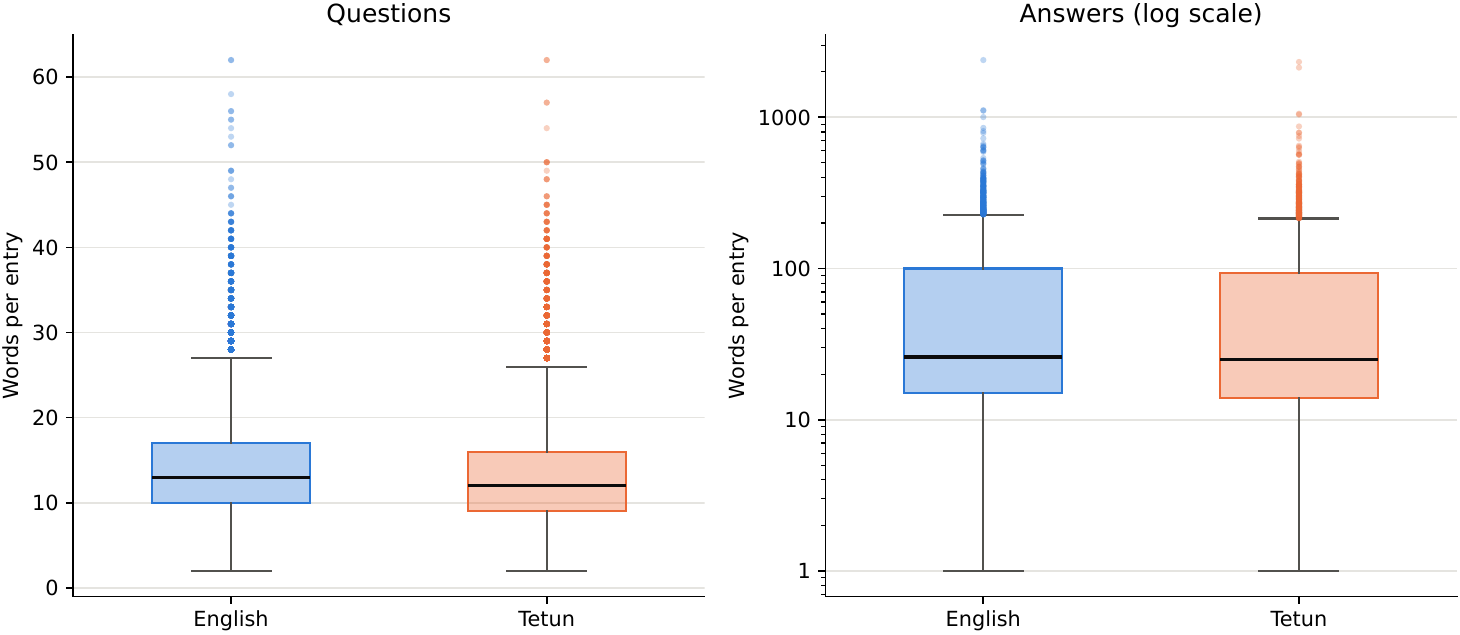}
\Description{Word-count distributions of English source texts and their Tetun translations. Question lengths are displayed on a linear scale, while answer lengths are displayed on a logarithmic scale.}
\caption{Word-count distributions of English source texts and their Tetun translations. Question lengths are displayed on a linear scale, while answer lengths are displayed on a logarithmic scale.}
\label{fig:word-distribution}
\end{figure}

We analyzed the word counts of English questions and answers and their Tetun translations. English questions contain an average of 13.74 words, compared with 13.33 words in Tetun. Similarly, English answers average 56.44 words, compared with 53.31 words in Tetun. Fig.~\ref{fig:word-distribution} shows similar question-length distributions across both languages, while answer lengths exhibit considerably greater variation, indicating that after translations, the words are reduced in Tetun when English documents are longer.

\section{Translation Quality Assessment}

To assess translation quality, we randomly selected 50 QA pairs (100 documents: 50 questions and 50 answers) from the two MedAlpaca-derived collections, with 25 pairs from each collection. We evaluated the LLM-generated translations using automatic metrics against human reference translations and assessments by two native Tetun-speaking healthcare professionals, focusing on medical accuracy and Tetun fluency.

The 100 selected English documents were provided to a native Tetun-speaking professional translator to produce human reference translations. Each question and answer was translated from English into Tetun while preserving the medical meaning of the original content. The resulting human translations served as reference texts for automatic evaluation and as a comparison set for assessing differences in medical accuracy and Tetun fluency between human and LLM-generated translations.

We evaluated the LLM-generated translations against the human reference translations using three automatic metrics: BLEU-4~\cite{papineni-et-al-2002-bleu}, chrF++~\cite{popovic-2017-chrf}, and ROUGE-L~\cite{lin-2004-rouge}. These metrics capture complementary aspects of translation quality. BLEU-4 measures modified n-gram precision up to four-grams, chrF++ combines character-level and word-level n-gram matching, and ROUGE-L measures sequence similarity based on the longest common subsequence.
All three metrics were computed using the same 100-document evaluation sample. We also examined the effect of orthographic normalization on the evaluation scores. Two settings were considered: (1)~the original Tetun texts without additional normalization and (2)~normalized texts in which hyphens were replaced with spaces, and apostrophes and diacritics were removed.

Given that automatic metrics cannot directly establish whether medical information has been accurately preserved, we conducted a human evaluation to assess medical accuracy and Tetun fluency in both human- and LLM-generated translations. Two native Tetun-speaking healthcare professionals evaluated the translations without knowing whether each translation was produced by a human translator or an LLM.

Translation quality was assessed using two criteria: (1)~\textit{Medical accuracy:} Whether clinical or health information remains correct and preserves the intended medical meaning; and (2)~\textit{Tetun fluency:} Whether the translation is natural, clear, and understandable in Tetun.
Each criterion was rated on a five-point scale, ranging from 1 (very poor) to 5 (excellent). We used Label Studio\footnote{\url{https://labelstud.io/}} to manage the annotation process. Inter-annotator agreement was measured using quadratic-weighted Gwet's AC2~\cite{gwet-2021}, which accounts for the ordinal nature of the ratings. Differences between human- and LLM-generated translations were assessed using the one-sided paired Wilcoxon signed-rank test~\cite{wilcoxon-1945}, based on the mean ratings of the two annotators for each document.

\section{Fine-Tuning Setup and Downstream Evaluation}

To investigate the downstream utility of \textit{Labadain-MedQA}, we used the translated medical QA data to fine-tune open-weight LLMs through PEFT. We selected six models from the Gemma and LLaMA families: Gemma 4 (E2B, E4B, and 12B), LLaMA 3.1 (8B), and LLaMA 3.2 (1B and 3B). Model selection was informed by previous Tetun-related studies~\cite{jesus-nunes-2026-labadainchat,jesus-nunes-2024-llm4eval} and considered different architectures, parameter sizes, and suitability for resource-constrained environments.

The 39,077 Tetun medical QA pairs from the MedAlpaca-derived component of \textit{Labadain-MedQA} were used for fine-tuning. Each question and its corresponding answer were formatted as an instruction-response example according to the input requirements of each model. The dataset was randomly divided into training (90\%), validation (5\%), and test (5\%) sets.
We applied QLoRA using the Unsloth framework~\cite{han-han-2023-unsloth}. The base models were quantized using 4-bit NormalFloat (NF4), and LoRA adapters were applied to all attention and multilayer perceptron (MLP) modules.
We used a LoRA rank ($r$) of 16, a scaling parameter ($\alpha$) of 16, zero dropout, and no bias adaptation.

All models were fine-tuned for three epochs with a maximum sequence length of 1,024 tokens, a per-device batch size of 2, and 16 gradient accumulation steps. We used a learning rate of $2 \times 10^{-5}$, the 8-bit AdamW optimizer, a weight decay of 0.001, and a linear learning rate scheduler with a warm-up ratio of 0.03. Model checkpoints and evaluation were performed every 250 optimization steps, and the checkpoint with the lowest validation loss was selected for downstream evaluation. All experiments were conducted on a single NVIDIA GeForce RTX 4090 GPU (24 GB VRAM).


We evaluated the fine-tuned models on 325 Tetun-translated USMLE multiple-choice questions covering Steps 1, 2, and 3 to assess their medical knowledge, clinical reasoning, and ability to apply medical knowledge in clinical scenarios. The experiments compared performance across the Gemma and LLaMA families and assessed the effects of fine-tuning by comparing the best-performing model against its corresponding base model. All models were evaluated using the same questions and prompts, with instructions to select the most appropriate answer from the available options.

Performance was measured using accuracy, defined as the proportion of questions for which the selected answer matched the reference answer.
For the best-performing model, we additionally evaluated two inference configurations using the trained QLoRA adapters: (1)~NF4 inference with the adapters applied to the quantized base model and (2)~BF16 inference after merging the adapters into the original BF16 base model. The corresponding base models were also evaluated in NF4 and BF16 precision to assess the effects of fine-tuning under each setting.

\section{Results and Discussion}

This section presents the results of human and automatic translation quality assessments and the downstream evaluation of fine-tuned models on Tetun medical QA tasks.

\subsection{Human Translation Quality}

\begin{figure}
\setlength{\fboxsep}{0pt}
\setlength{\fboxrule}{0.2pt}
\fcolorbox{gray!50}{white}{
    \includegraphics[width=0.96\textwidth]{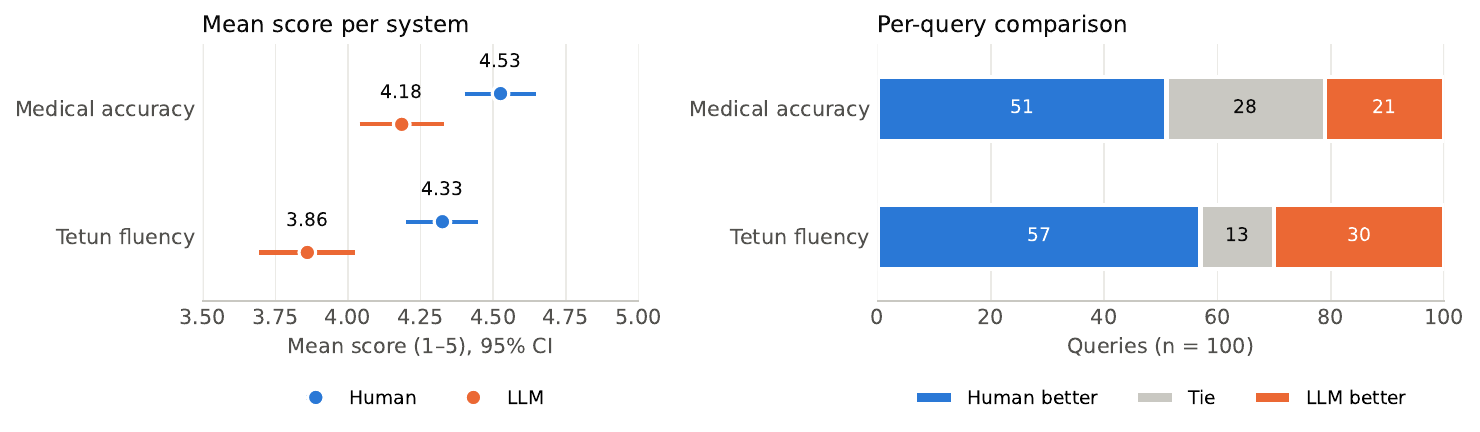}
}
\Description{Human evaluation of medical accuracy and Tetun fluency for human and LLM translations, showing mean ratings with 95\% confidence intervals and per-document comparisons ($n=100$).}
\caption{Human evaluation of medical accuracy and Tetun fluency, showing mean ratings with 95\% confidence intervals and per-document comparisons ($n=100$).}
\label{fig:human_eval_a_b_systems}
\end{figure}

Fig.~\ref{fig:human_eval_a_b_systems} presents the human evaluation results for medical accuracy and Tetun fluency. Human translations achieved higher mean ratings than LLM-generated translations for both criteria. Medical accuracy averaged 4.53 for human translations and 4.18 for LLM translations, while Tetun fluency averaged 4.33 and 3.86, respectively. The overall mean ratings were 4.43 for human translations and 4.02 for LLM translations on a five-point scale.
The per-document comparison shows that human translations received higher ratings for both medical accuracy and Tetun fluency. 
Overall, both human and LLM-generated translations achieved high ratings, although human translations performed significantly better. These findings suggest that automatic translation is a promising approach for developing Tetun medical QA resources.

Table~\ref{tab:translation_statistical_tests} presents the statistical comparison between human and automatic translations. Both the Wilcoxon signed-rank and paired $t$-tests showed statistically significant differences in medical accuracy and Tetun fluency (all $p < 0.001$). Kendall's $\tau_b$~\cite{kendall-1945} coefficients were 0.231 for medical accuracy and 0.254 for Tetun fluency, indicating weak positive associations between human and LLM translation ratings across documents.

\begin{table}
\caption{Statistical comparison of human and LLM-generated translations for medical accuracy and Tetun fluency. Values are reported as mean $\pm$ STD of per-document annotator means ($n=100$).}
\label{tab:translation_statistical_tests}
\centering
\small
\setlength{\tabcolsep}{4pt}
\begin{tabular}{lccccc}
\hline
\textbf{Criterion} & \textbf{Human} & \textbf{LLM} &
\textbf{Kendall's $\tau_b$} & \textbf{Wilcoxon $p$} &
\textbf{Paired $t$-test $p$} \\
\hline
Medical accuracy
& $4.53 \pm 0.63$
& $4.18 \pm 0.74$
& 0.231
& $1.0 \times 10^{-4}$ 
& $1.4 \times 10^{-4}$ \\

Tetun fluency
& $4.33 \pm 0.63$
& $3.86 \pm 0.84$
& 0.254
& $7.4 \times 10^{-5}$
& $3.9 \times 10^{-5}$ \\
\hline
\end{tabular}
\end{table}

Inter-annotator agreement was measured using quadratic-weighted Gwet's AC2. Human translations achieved agreement coefficients of 0.8640 for medical accuracy and 0.8070 for Tetun fluency. The corresponding coefficients for LLM translations were 0.7290 and 0.7070. These results indicate substantial to almost perfect agreement between the two healthcare professionals, with higher agreement for human translations.

\subsection{Automatic Translation Quality}

The automatic evaluation results using BLEU-4, chrF++, and ROUGE-L are presented in Table~\ref{tab:automatic_eval_results}. Orthographic normalization, which involved replacing hyphens with spaces and removing apostrophes and accents, improved all three metrics, with gains of 3.23 points for BLEU-4, 2.36 points for chrF++, and 2.52 points for ROUGE-L.

\begin{table}
\caption{Automatic evaluation of English-to-Tetun medical QA translations before and after orthographic normalization. Improvements indicate absolute score differences.}
\label{tab:automatic_eval_results}
\centering
\begin{tabular}{lrrr}
\hline
\textbf{Method} & \textbf{BLEU-4} & \textbf{chrF++} & \textbf{ROUGE-L} \\
\hline
Without normalization & 23.45 & 55.17 & 58.78 \\
With normalization    & 26.68 & 57.53 & 61.30 \\
\hline
\textit{Improvement}  & \textit{+3.23} & \textit{+2.36} & \textit{+2.52} \\
\hline
\end{tabular}
\end{table}

These improvements highlight the sensitivity of automatic evaluation metrics to Tetun orthographic variation. Differences involving hyphens, apostrophes, and diacritics can affect lexical matching even when the underlying meaning remains unchanged. However, higher scores after normalization do not necessarily indicate improved medical accuracy, as normalization only affects the text representation.

\subsection{Downstream Model Performance}

Table~\ref{tab:usmle_qa_accuracy} presents the downstream evaluation results for the Gemma and LLaMA model variants on 325 Tetun-translated USMLE multiple-choice questions. Accuracy was measured separately for Steps 1, 2, and 3, along with the average across the three steps.
Among the QLoRA-fine-tuned models evaluated with NF4 inference, Gemma 4 12B achieved the highest mean accuracy of 0.5389, followed by Gemma 4 E4B (0.4595) and E2B (0.2625). All LLaMA variants achieved lower mean accuracies than the Gemma models, suggesting that model architecture, size, and pretrained capabilities may influence downstream performance.

\begin{table}
\caption{Accuracy of QLoRA-fine-tuned Gemma and LLaMA model variants on Tetun-translated USMLE multiple-choice questions across Steps 1--3. \underline{Underlined} values indicate the best performance under NF4 inference, while \textbf{bold} values indicate the best overall performance.}
\label{tab:usmle_qa_accuracy}
\centering
\begin{tabular}{|l|r|r|r|r|}
\hline
\textbf{Model} & \textbf{Step 1} & \textbf{Step 2} & \textbf{Step 3} & \textbf{Average} \\
\hline
Gemma 4 E2B + QLoRA (NF4) & 0.2766 & 0.2569 & 0.2541 & 0.2625 \\
Gemma 4 E4B + QLoRA (NF4) & 0.4362 & 0.4587 & 0.4836 & 0.4595 \\

LLaMA 3.1 8B + QLoRA (NF4) & 0.2447 & 0.2110 & 0.1967 & 0.2175 \\
LLaMA 3.2 1B + QLoRA (NF4) & 0.1596 & 0.2294 & 0.1393 & 0.1761 \\
LLaMA 3.2 3B + QLoRA (NF4) & 0.1596 & 0.2477 & 0.2049 & 0.2041 \\

Gemma 4 12B Base (NF4) & 0.5106 & 0.4404 & 0.5082 & 0.4864 \\
Gemma 4 12B + QLoRA (NF4) & \underline{0.5638} & \underline{0.4954} & \underline{0.5574} & \underline{0.5389} \\

Gemma 4 12B Base (BF16) & 0.5213 & 0.5688 & 0.5656 & 0.5519 \\
Gemma 4 12B + QLoRA (BF16, Merged) & \textbf{0.6170} & \textbf{0.5963} & \textbf{0.6148} & \textbf{0.6094} \\
\hline
\end{tabular}
\end{table}

For Gemma 4 12B, QLoRA fine-tuning improved mean accuracy from 0.4864 to 0.5389 under NF4 inference, representing a relative gain of 10.79\%. However, the fine-tuned NF4 model remained slightly below the BF16 base model (0.5519), suggesting that fine-tuning did not fully compensate for the performance differences associated with quantization. When the trained LoRA adapters were merged into the BF16 base model, mean accuracy increased to 0.6094, representing a relative gain of approximately 10.42\% over the BF16 base model. These findings highlight the importance of considering both fine-tuning and inference precision when evaluating downstream performance.

The BF16 merged model achieved the highest accuracy across all three USMLE steps: 0.6170 for Step 1, 0.5963 for Step 2, and 0.6148 for Step 3. Compared with the BF16 base model, the largest relative improvement occurred in Step 1 (+18.36\%), followed by Step 3 (+8.70\%) and Step 2 (+4.83\%). These differences suggest that the benefits of fine-tuning may vary depending on the characteristics of the medical questions. Overall, the results provide encouraging evidence that translated Tetun medical QA data can improve downstream model performance.

\subsection{Discussion}

The findings suggest that automatic translation is a promising approach for developing medical QA resources in Tetun. Although human translations achieved significantly higher ratings, LLM-generated translations received an overall mean rating of 4.02 out of 5, indicating generally high quality. The larger difference in Tetun fluency than in medical accuracy suggests that producing natural Tetun remains an important challenge. Moreover, the variation in per-document ratings highlights the need for quality control, particularly in medical contexts where individual translation errors may affect the interpretation of health information.

The improvements in automatic evaluation scores after orthographic normalization further highlight the challenges of evaluating low-resource languages. Differences in spelling conventions can substantially influence lexical similarity metrics without necessarily affecting the underlying meaning. This suggests that automatic metrics should be interpreted alongside human assessments, particularly when evaluating medical content.

The downstream results provide further evidence of the utility of automatically translated medical QA data. Fine-tuning improved Gemma 4 12B performance under both NF4 and BF16 inference, suggesting that translated medical knowledge can contribute to downstream tasks beyond the fine-tuning dataset. However, the differences across model families are particularly notable. Gemma 4 E4B substantially outperformed the larger LLaMA 3.1 8B model, indicating that parameter size alone is not a reliable predictor of performance. Differences in model architecture, pretraining, and language representation may also influence how effectively models learn from translated data.

Another important finding concerns the relationship between fine-tuning and inference precision. Although QLoRA improved the NF4 Gemma 4 12B model, its accuracy remained below that of the original BF16 base model. The highest performance was achieved when the trained LoRA adapters were merged into the BF16 base model. This suggests that the benefits of fine-tuning can be influenced by the inference configuration and that improvements over a quantized baseline do not necessarily translate into improvements over a higher-precision base model. The variation in relative gains across USMLE steps also suggests that adaptation benefits may differ across medical question types, although further analysis is needed to explain these differences.

\section{Conclusion and Future Work}

This paper introduced \textit{Labadain-MedQA}, a Tetun medical QA resource developed through automatic translation and evaluated through human assessments and downstream LLM fine-tuning. The results indicate high translation quality and improvements in downstream performance, particularly for Gemma 4 12B. Differences across model families and inference configurations further highlight the importance of model selection and inference precision. These findings suggest the potential of translated medical QA resources for supporting under-resourced language model adaptation.

Future work will focus on improving translation quality through expert linguistic review, particularly Tetun grammatical correctness, orthographic consistency, and terminology. We also plan to expand evaluations to additional models and medical tasks, assess medical reasoning, factual accuracy, and reliability, and extend the resources to support broader research on Tetun medical language technologies.

\section{Resource Availability and Limitations}

The \textit{Labadain-MedQA} datasets, models, and evaluation resources publicly are accessible at \url{https://github.com/labadain/labadain-medqa}, released under the Creative Commons Attribution-ShareAlike (CC BY-SA) license.
The resources include 39,077 translated medical QA pairs, 325 Tetun-translated USMLE-based questions, a human-translated reference set of 100 medical QA documents, human evaluation annotations assessing medical accuracy and Tetun fluency for both human and LLM-generated translations, and six fine-tuned models from the Gemma 4 and LLaMA 3 families.

This study has several limitations. Translation quality was assessed on a sample of 100 documents, which may not fully represent the entire dataset, while errors in automatic translations may affect both training and evaluation. Direct comparisons with base models were limited to Gemma 4 12B, and performance on medical multiple-choice questions alone does not establish clinical reliability. Given these limitations and the potential patient-safety concerns, the models are intended for research purposes and require expert validation and safety evaluation before use in healthcare applications.

%
%
%
%

\end{document}